# Generative AI–Assisted Workflows in Architectural Conceptual Design: Performance, Creative Self-Efficacy, and Cognitive Load

Yao Xiao*, Han Jiang*, Rachel Hurley, Shichao Liu†

*Department of Civil, Environmental, and Architectural Engineering, Worcester Polytechnic Institute, Worcester, MA, US*

*These authors contributed equally to this work.

*sliu8@wpi.edu*

## Abstract

Generative AI (GenAI) is increasingly adopted in design education, yet evaluating its educational value through final outcomes provides an incomplete picture. This study compares two ecologically plausible workflows in an architectural conceptual design task: GenAI-assisted image generation and ArchDaily-based precedent search. The comparison concerns complete workflows rather than the isolated contributions. Thirty-six students completed a two-phase design task, first designing independently and then revising with their assigned workflow. Screen-recording demonstrated active engagement in both conditions. Eight judges rated design performance, while participants reported task-specific and general creative self-efficacy and cognitive load after each phase. Difference-in-differences analyses showed no significant overall differences between the GenAI and precedent-search workflows in design performance, cognitive workload, or task-specific creative self-efficacy. Beyond these null overall effects, three patterns were observed. General creative self-efficacy showed a significant relative decline under the GenAI workflow. A subgroup analysis suggested higher revision-phase performance among novice students using GenAI than among those using precedent search ($F(1,32) = 4.303$, $p = 0.046$, partial $\eta^2 = 0.118$). However, this exploratory interaction should be interpreted cautiously due to low rating reliability, small subgroup cells, and imprecise estimation. Third, exploratory prompt analyses suggested that iterative, task-specific prompting strategies (CD3, CD6) were associated with cognitive load reductions at the uncorrected level, but neither association survived multiple-comparison correction. Overall, the GenAI workflow did not produce uniform gains. Its educational value may depend on pedagogical framing, learner characteristics, and human–AI interaction structure, underscoring the need to preserve creative agency and develop prompt literacy.

## Nomenclature

| | |
|---|---|
| CAD | Computer-Aided Design |
| CLT | Cognitive load theory |
| DiD | Difference-in-Differences |
| GenAI | Generative Artificial Intelligence |
| GPT | Generative Pre-trained Transformer |
| SD | Standard deviation |
| STEM | Science, technology, engineering, and mathematics |

**Highlights**

- GenAI and search workflows showed no overall performance or cognitive-load gain.
- General creative self-efficacy declined significantly under the GenAI workflow.
- An exploratory subgroup analysis suggested a preliminary novice benefit.
- Iterative prompt patterns linked to cognitive-load reduction before correction.

## 1. Introduction

Architectural conceptual design is a visually demanding activity that requires students to develop a cluster of interrelated skills: spatial reasoning, massing exploration, and the visual communication of design intent (the translation of abstract programmatic goals into legible sketches, diagrams, and spatial representations that convey meaning to diverse audiences) (Friel et al., 2001; Hegarty, 2011). Effective visual communication requires clarity, legibility, and readability, incorporates an appropriate level of detail and complexity to represent relationships and concepts (Pei et al., 2026; Stenliden & Nissen, 2022), and ultimately serves to convey design intent and support shared understanding among diverse audiences (Chandrasegaran et al., 2018; Downes & Lange, 2015). Together, these skills constitute the foundation of what designers call ideation: the generative, iterative process of producing and evaluating emerging design ideas through visual means. Ideation is distinct from final design production, as it is the exploratory phase in which designers externalize tentative concepts, compare alternatives, and progressively clarify design direction before committing to detailed resolution (Nakakoji et al., 1999; Dinar et al., 2015; Arbib, 2018; Wijetunge et al., 2025). For students, especially those without formal training in sketching or visual ideation, these demands can constrain creative exploration, weaken visual communication, and increase workload (Seneviratne et al., 2022; Martín Erro et al., 2024; Fareed et al., 2024).

Recent advances in generative artificial intelligence (GenAI) have made it possible to generate visual content rapidly from natural-language prompts in conversational text-to-image interface (Alier et al., 2024; Bian et al., 2025). In architecture and design more broadly, GenAI has been increasingly used to visualize concepts, explore alternatives, and accelerate idea development during early design stages (C. Li et al., 2025; Royal Institute of British Architects (RIBA), 2024). This emerging use has drawn increasing attention to how AI-generated images affect students' learning outcomes and processes, as well as to their pedagogical implications for teaching and learning.

As prior studies have shown, GenAI assistance can improve productivity and performance in tasks such as writing and programming (Al-Raimi et al., 2024; Daniotti et al., 2026; Pellas, 2023). However, evidence remains limited in architectural conceptual design, especially in visual ideation tasks that require students to iteratively interpret, evaluate, and revise generated outputs. Moreover, the existing literature in design education has tended to focus on design outcomes, efficiency, or general perceptions of AI use, rather than on the learner-centered processes through which GenAI may influence design activity (Bian et al., 2025; Seneviratne et al., 2022; Urban et al., 2025). In this study, we therefore focus on two learner-centered constructs beyond design outcomes: cognitive load and creative self-efficacy. Cognitive load captures the mental effort required to process information and manage task demands during design work (Lavrsen & Daalhuizen, 2024), while creative self-efficacy reflects students' beliefs in their ability to produce creative outcomes (Bandura & Wessels, 1997; Haase et al., 2018). Recent review studies suggest that GenAI can support idea generation and problem solving under some conditions, but its effects are highly contingent on the nature of the task, the scaffolding provided, and the ways in which learners use the tool (C. Li et al., 2026; Vendrell & Johnston, 2026).

Across studies, findings remain mixed: some report reduced perceived load, likely because AI offloads part of the task demand (Chandrasekera et al., 2025; C.-H. Chen & Chang, 2024), whereas studies using electroencephalogram (EEG) measures have found comparable or even higher cognitive load during cognitively demanding tasks (Schulz & Knierim, 2024). Other studies raise concerns about cognitive offloading, reduced critical engagement, and weakened creativity when learners rely on AI outputs without sufficiently evaluating or transforming them (K.-L. Huang et al., 2024; Fu et al., 2026; Dwivedi et al., 2023; Peters et al., 2024). At the same time, research also suggests that when GenAI is used as an augmentative tool rather than a replacement, it can support idea development and improve performance in creative problem-solving tasks beyond either AI-only or human-only approaches (Fan et al., 2025; Hitsuwari et al., 2023; Kim & Lee, 2023; Lin et al., 2020).

Furthermore, novice and experienced learners engage with learning tools differently due to variations in domain-specific prior knowledge, self-efficacy, and technology acceptance (K.-L. Huang et al., 2024), as well as how students frame prompts, interpret AI outputs, and decide when to iterate versus rely on their own reasoning, substantially shape both interaction patterns and academic outcomes (Urban et al., 2025). A recent study found that experienced learner demonstrated higher quality, novelty, and creativity when engaging with AI chatbots (Sun et al., 2022). Research has suggested greater benefits for novice users of ChatGPT in writing tasks (Doshi & Hauser, 2024; Noy & Zhang, 2023). Therefore, the broader impact of GenAI on different levels of group in visual design field remains largely unknown.

As a result, these findings suggest that GenAI does not exert a uniform effect; rather, task demands, learner characteristics, and interaction strategies may jointly shape both learning processes and outcomes. Accordingly, the educational value of GenAI in architectural conceptual design cannot be inferred from output quality alone, but must also be examined through variables that capture how students experience and manage the design process. To address this gap, we conducted a controlled two-phase architectural conceptual design experiment with 36 students. Participants first completed an initial design independently and then revised their work using either GenAI-assisted image generation with DALL-E 3 in ChatGPT-4 or ArchDaily-based precedent search. These conditions represented two ecologically plausible but structurally different visual-support workflows. We evaluated design performance through expert ratings, tracked phase-based changes in cognitive workload and task-specific and general creative self-efficacy, and analyzed screen recordings to examine how participants interacted with the assigned workflows. We further investigated whether workflow-related differences varied by initial competency and how prompt-use patterns were associated with changes in cognitive workload. In doing so, this study moves beyond asking whether GenAI improves design outputs to instead investigate the dynamics of human–AI interaction during ideation, with the goal of offering a more educationally meaningful account of GenAI in architectural conceptual design learning.

Based on this framing, the study addresses the following questions:

**RQ1.** How does students' design performance, cognitive load, and creative self-efficacy differ between the GenAI-assisted and precedent-search workflows in an architectural conceptual design task?

**RQ2.** Do GenAI-assisted workflow vary across students with different initial competency levels?

**RQ3.** How are different prompt-use patterns associated with changes in learners' cognitive load during design revision?

## 2. Related Work

### 2.1 Image Generation with GenAI for Architectural Design

Recent advances in text-to-image generation have been led by models such as Midjourney, DALL-E 3, and Stable Diffusion, each offering different affordances for design-related tasks rather than forming a simple hierarchy of quality. Midjourney has often been noted for producing highly stylized and visually polished images, making it useful for rapid aesthetic exploration, although its closed platform and limited controllability can constrain precise design iteration. DALL-E 3, integrated with GPT-based prompt interpretation, supports close translation from natural-language design intentions to images within a conversational workflow, but its outputs may still contain representational biases and may not reliably satisfy domain-specific architectural constraints (Currie et al., 2025; Geng & Yang, 2025; Liao et al., 2024). Stable Diffusion offers greater openness, customization, and controllability, especially when combined with tools such as ControlNet or LoRA, although achieving consistent high-quality outputs often requires more technical configuration and model tuning (S.-Y. Jang & Kim, 2024; Pei et al., 2026; Ploennigs & Berger, 2023). Thus, the suitability of each system depends on the design task, desired level of control, workflow constraints, and user expertise.

Text-to-image GenAI is becoming a normal, expected part of architectural education, especially in early design stages like ideation and visualization (Dullinja & Jashanica, 2025). By converting textual or verbal prompts into visually detailed outputs, these systems enable rapid conceptual exploration and visualization, accelerating the generation of high-quality design representations and facilitating more iterative, exploratory workflows (Abrusci et al., 2025; Lystbæk, 2025). Rather than treating image generation only as a tool for producing polished outputs, recent work suggests that GenAI is often used during early ideation to help designers externalize tentative concepts, generate alternatives, and iteratively refine visual directions in response to evolving design intent (C. Li et al., 2025). Beyond productivity, text-to-image GenAI supports innovative and user-centered design thinking by enabling architects to rapidly experiment with alternative styles during the sketching phase.

## 2.2 GenAI's Effects on Creative Self-Efficacy and Cognitive Load

Prior research shows that supportive creative environments can directly enhance students' creative self-efficacy and creative thinking (Zeng, 2025). GenAI contributes to this process by rapidly iterating and visualizing designers' ideas without demanding extensive time or technical skill, thereby facilitating early-stage ideation for students and novice designers. Self-efficacy plays a crucial moderating role in this interaction, with higher levels of confidence linked to more exploratory engagement and more effective use of AI tools (Hwang & Wu, 2025; Noy & Zhang, 2023). Survey data from Guo et al. further indicate that creative self-efficacy significantly moderates creativity gains in everyday and professional uses of GenAI (2025). In early computational design education, establishing students' confidence has been shown to have a stronger impact on creativity than merely increasing motivation. Complementing this, a recent meta-analysis of experimental studies found that ChatGPT interventions were associated with improved academic performance, affective-motivational states, and higher-order thinking propensities, reduced mental effort, and no statistically significant overall effect on self-efficacy, although these effects varied across contexts and modes of use (Deng et al., 2025).

Although fewer studies have examined these mechanisms in architectural design, emerging evidence suggests that GenAI-supported design instruction can meaningfully enhance students' creative self-efficacy. For example, Huang et al. (2024) integrated multimodal AI into architectural design workflows and found that AI-assisted strategies significantly outperformed traditional instruction in boosting self-efficacy, with more advanced students reporting higher confidence than novices. In elementary visual arts education, students using AI-generated imagery exhibited significantly higher self-efficacy than those under traditional teaching, as the visual appeal and specificity of AI outputs strengthened students' perceived competence in understanding and applying new material (Bian et al., 2025).

At the same time, GenAI introduces a distinctive cognitive profile in creative tasks. It can reduce cognitive load by offloading routine visual-ideational processes, yet it can also increase cognitive load when users must interpret ambiguous outputs, formulate effective prompts, or manage iterative refinements. Unlike traditional tools, GenAI redistributes cognitive effort from low-level manual production toward higher-level evaluation, judgment, and decision-making (Lee et al., 2025). In this sense, it can create a distinctive creative profile by shifting students' effort away from manual visual production and toward prompt formulation, output evaluation, and the selective integration of AI-generated alternatives. Consequently, the impact of AI on workload in creative design is not uniformly beneficial; it depends on learners' prior knowledge, prompting literacy, and the alignment between AI outputs and task goals.

High-quality prompting also appears to be a learned cognitive skill. In a study of non-routine problem solving, providing ChatGPT with explicit evaluation criteria, detailed contextual background, and iterative feedback substantially improved the originality and specificity of generated solutions. Higher-performing participants used more frequent and more detailed prompts, suggesting that effective prompting itself contributes to better design outcomes (Urban et al., 2025). Empirical findings in visual education further indicate that GenAI does not necessarily increase overall workload. In an elementary art study, students using GenAI showed no significant differences in cognitive load relative to traditional teaching, implying that AI-generated images did not add learning burden while helping maintain conceptual complexity through intuitive visual scaffolding (Bian et al., 2025). In higher education design contexts, Chandrasekera et al. (2025) found that students working on urban furniture concepts experienced significantly lower cognitive load when using AI compared with traditional sketching, and GenAI substantially enhanced lateral and abstract thinking. Similarly, in an ideation and concept-development experiment with trained designers, AI-assisted brainstorming significantly reduced perceived mental demand. Eye-tracking analyses showed shorter visual search paths and more distributed attention under AI assistance, indicating reduced extraneous cognitive load (Fu et al., 2026). Chen et al. (2025) compared a structured, choice-based interface (Choices2I) with the traditional text-based prompting used in GPT-4's official web version. They found that users from creative industries (e.g., designers, artists, film directors) consistently preferred Choices2I because it offered greater flexibility and control, regardless of their expertise level. In contrast, users from non-creative fields, who favored text-based prompting, which imposed lower cognitive load and avoided the pressure of making multiple design choices.

## 3. Methodology

### 3.1 Participants

We conducted an experiment in Spring 2024 to examine the impact of GenAI on visual communication in architectural concept design. We first recruited 41 participants through flyers and snowball sampling. We then randomly assigned them to the Control group and GenAI group. However, 5 participants did not complete the study. Thirty-six students were formally considered and randomly assigned to either the Control group or the GenAI group. The Control group consisted of 18 participants (8 females), while the GenAI group had 18 participants (9 females). The participants represented a range of academic disciplines and years of study since the goal of this study is to understand the effectiveness of GenAI across general disciplines beyond architecture for external validity. Table S1 in the Appendix describes the study major of each participant. Participants were compensated with a total payment of $50 (Amazon gift card) for their participation in this experiment. Additionally, the top three participants received an extra $100 bonus based on expert evaluation of their design work.

### 3.2 Experimental Design

The experiment consisted of two visits: a group tutorial session (Visit 1) and an individual design session (Visit 2). Visit 1 provided participants with a shared baseline understanding of the architectural conceptual design task, as they came from diverse engineering backgrounds and had varying levels of design experience. Visit 2 was the formal, time-bounded individual design session.

During Visit 2, participants completed the task in two phases. In Phase 1, all participants developed an initial conceptual design without external visual-support tools. In Phase 2, participants revised their initial design using the assigned source of visual inspiration: participants in the Control group used ArchDaily.com, a popular online repository of real-world architectural projects, whereas participants in the GenAI group used DALL-E 3 within ChatGPT-4 to generate design images through prompting. Accordingly, the terms Control group and GenAI group refer to the ArchDaily-based precedent-search workflow and the DALL-E 3–assisted image-generation workflow, respectively. The design supports comparison of the assigned workflows as wholes but does not separately identify the contribution of any individual component. This two-phase structure allowed us to capture participants' self-generated concepts before tool use and then examine how different forms of visual support influenced the revision of an existing design direction.

The Control condition was based on precedent study, a common architectural design practice in which designers examine existing projects to inform new design decisions. Preliminary inquiry with architectural design instructors at WPI indicated that ArchDaily was commonly used for precedent-based ideation. It was therefore selected as an ecologically plausible precedent-search resource. To reduce time lost to ineffective web searching, particularly among participants without formal architectural training, the condition was restricted to this single repository rather than unrestricted internet browsing.

The GenAI condition used DALL-E 3 within ChatGPT-4 because the platform was widely accessible at the time of the study and provided an integrated conversational environment for formulating prompts, inspecting generated images, and iteratively refining requests. This configuration supported the study's focus on prompt-based interaction and conceptual design revision rather than comparison among image-generation systems. Despite their structural differences, both workflows served the same role in the revision task: participants examined visual material, evaluated its relevance to the design brief, and selectively incorporated it into their existing design direction. Initial and revised designs were evaluated by a panel of judges, and surveys assessed creative self-efficacy and perceived workload. Further details are provided below.

### 3.3 Architectural Conceptual Design Task

Participants were asked to develop an architectural concept design for a research laboratory building located at Gateway Park of WPI (Fig. 1). Gateway Park is a mixed-use facility housing a range of academic, research, and commercial enterprises. The site is located in the ASHRAE climate zone 5A. The design task was situated at a site near the WPI campus, which provided participants with a familiar physical context. In addition, the proposed laboratory building reflected a realistic project scenario, as a similar development had been proposed by a local development group (Worcester Business Journal, 2022). The task focused on the conceptual design of building massing and geometry rather than detailed technical resolution.
Given the diversity of participants' backgrounds, we gave a mini crash course on how to develop architectural design concepts in a separate visit before the formal experiment. The formal design experiment consisted of two phases including free design phase without consulting any design project repositories for precedent studies or applying any AI tools. The deliverable of each phase of the design task is a sketch using a 15.6" digital drawing tablet (Artist 15.6 Pro, XP-PEN) with a stylus. Fig. 1b displays sketches of the two phases from two participants in control and GenAI groups, respectively.

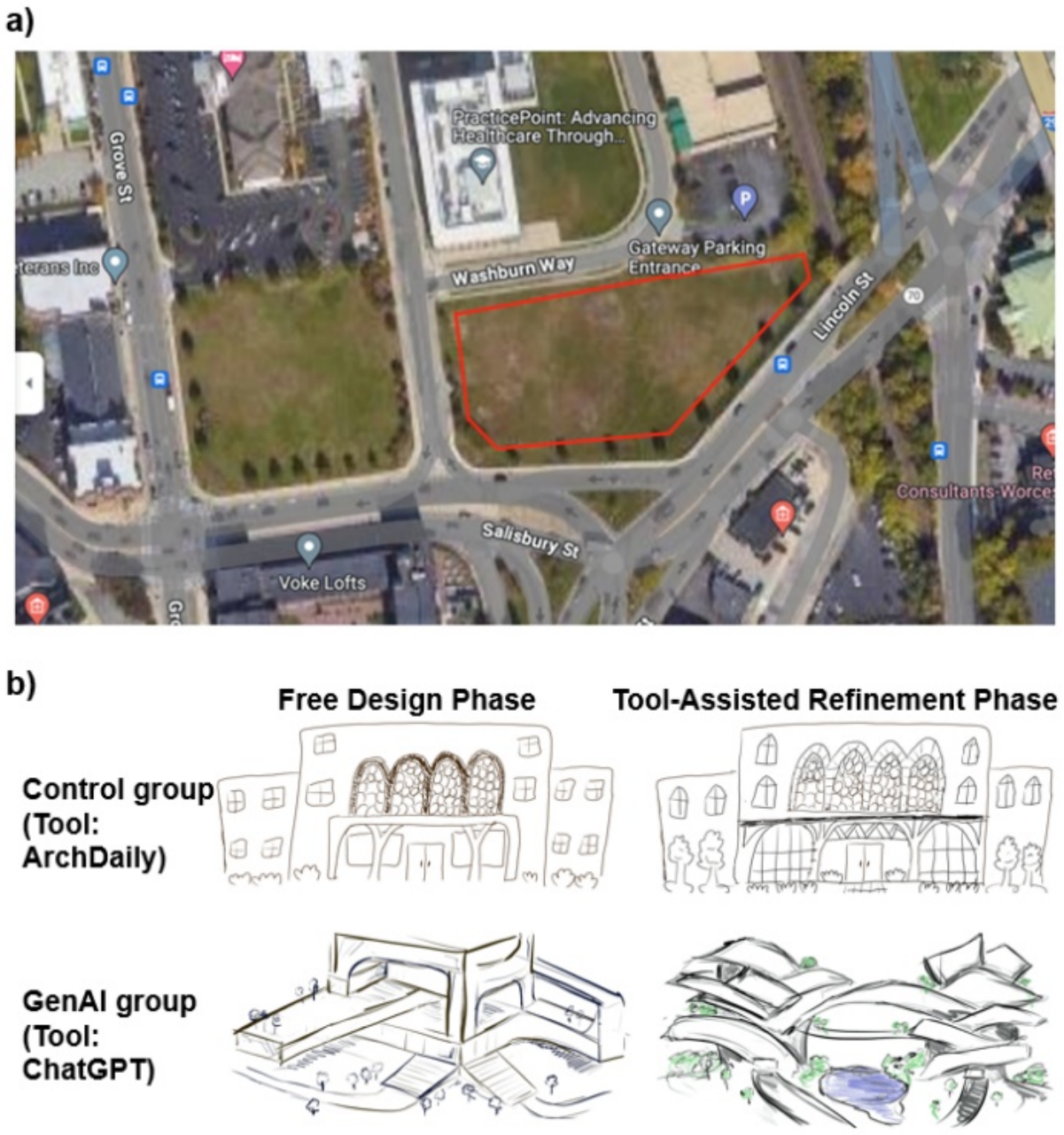

***Fig. 1.*** *Site and sample conceptual design; a) Design site and neighborhood of the research laboratory building; b) Design samples of two participants in Control and GenAI groups respectively.*

### 3.4 Procedure

Before formal data collection, three internal feasibility-test sessions were conducted to check the smoothness of the experimental workflow, including the online task interface, timing, participant instructions, drawing-tablet setup, and screen-recording procedure. Data from these sessions were not included in the final sample and were not analyzed.

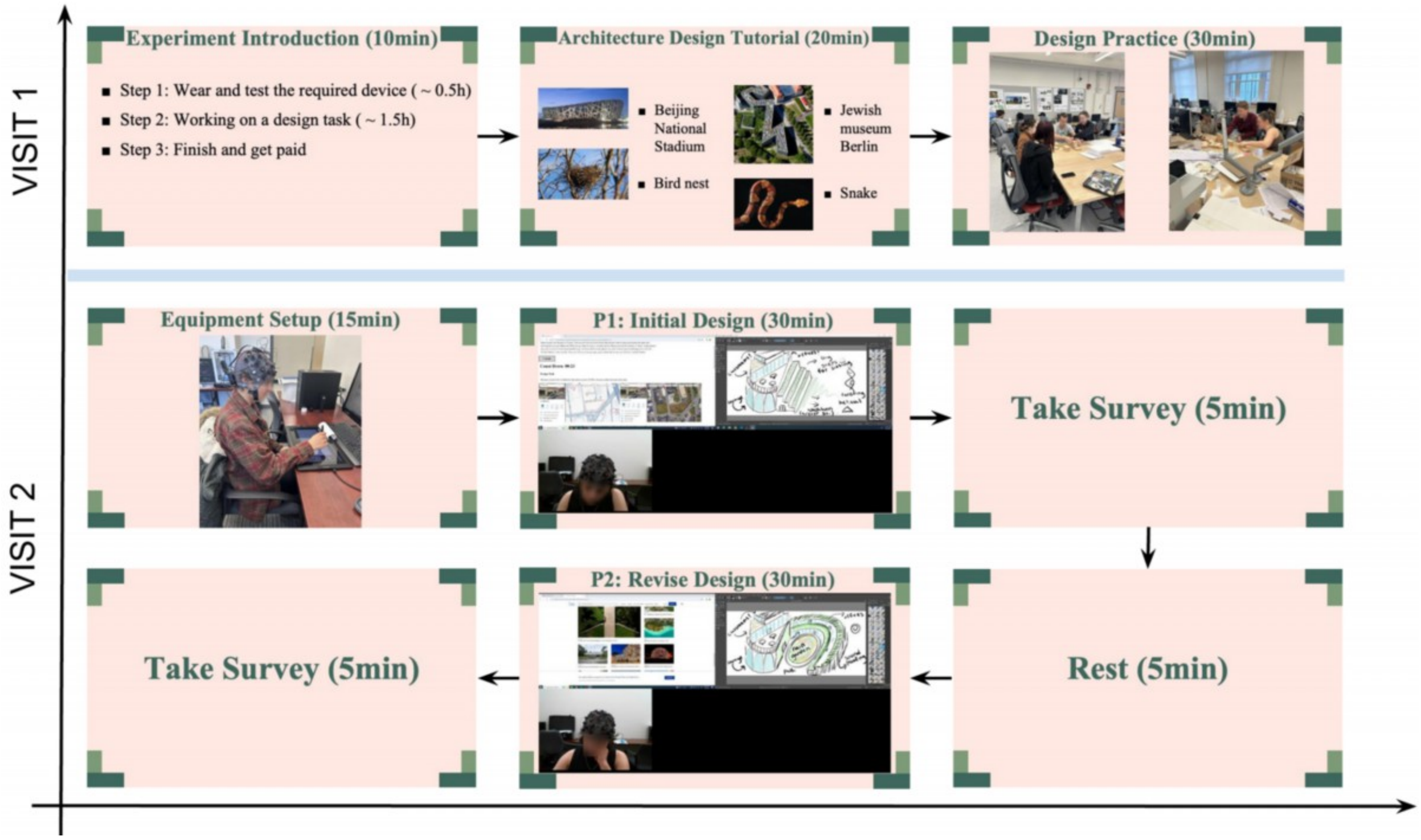

***Fig. 2.*** *Experimental protocol across two visits. Visit 1: Group tutorial; Visit 2: Individual design.*

### 3.4.1 Group Tutorial Session (Visit 1)

The first visit lasted approximately one hour. During this session, a professional provided a detailed explanation of the experiment, including an overview of the *design* task for the second visit. A brief lecture on fundamental design principles was also *provided*. To help participants familiarize themselves with the design process, *p*articipants *were instructed to* use these woodblocks to explore architectural concepts and practice design strategies *(**Fig.** 3)*. Both the Control and GenAI groups received the same tutorial content prior to random assignment in the individual session.

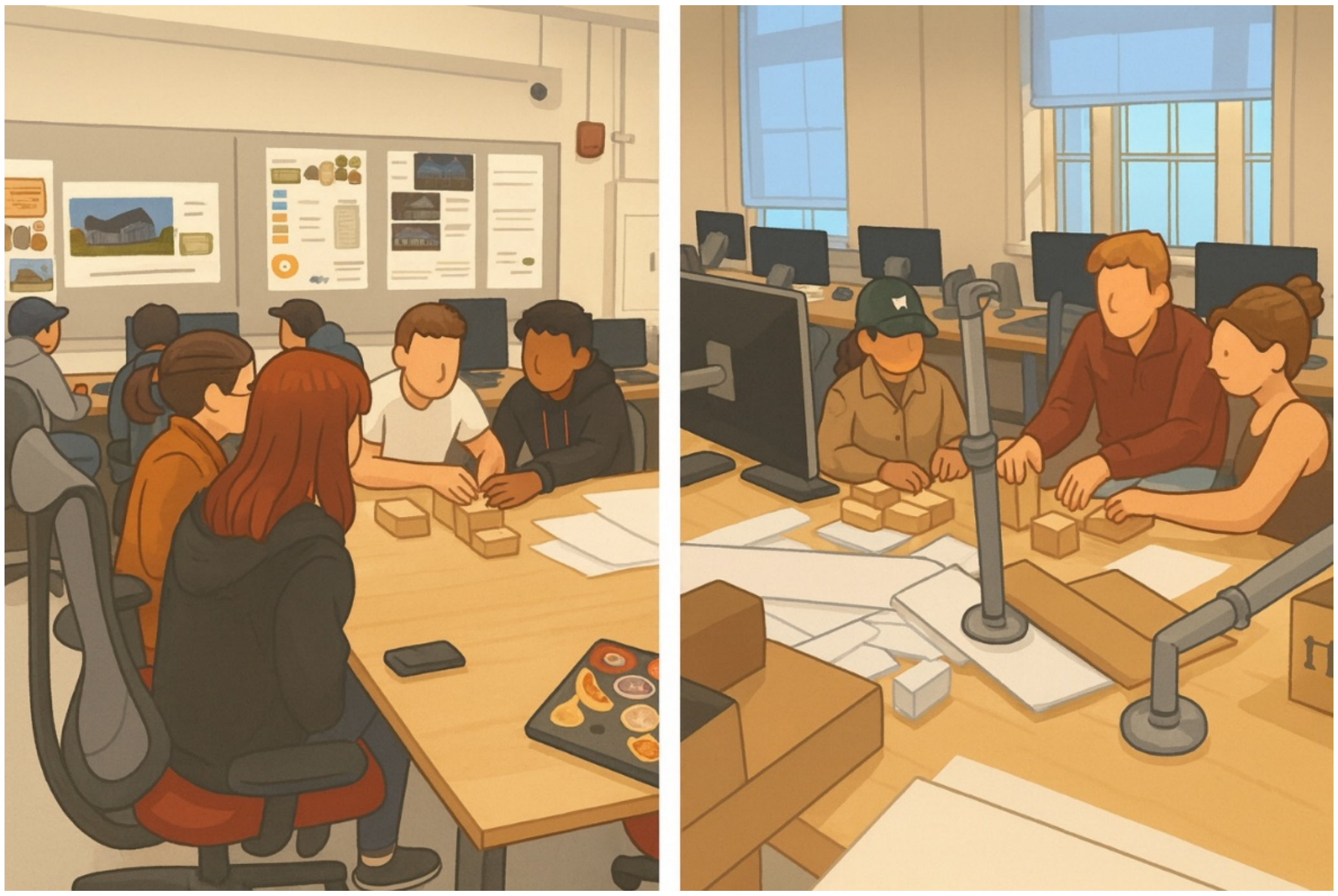

*Fig. 3. Students in the group tutorial used woodblock to learn conceptual design. The image was converted from a photo taken during a session to showcase the study setup while protecting participant privacy.*

#### 3.4.2 Individual Design Session (Visit 2)

The second visit lasted approximately 1.5 hours (**Fig. 2**). Each participant scheduled a time slot and was randomly assigned to either the control group or the treatment group. After Visit 2, each of the 36 participants will submit both an initial design drawing and a revised design drawing for their design concepts. The session was structured into three phases:

1) ***Setup Phase (approximately 30 min):*** A researcher assisted each participant with donning the neurophysiological sensors and configuring the software used for data collection. Details of the neurophysiological measures will be reported in future publications. After the setup was completed, the researcher left the lab, and the participant completed the remainder of the session independently. Step-by-step guidance was provided through an online platform that displayed dynamic countdowns indicating the remaining time for each task.

2) ***Initial Design Phase (30 min):*** Participants created their initial design on the drawing tablet using only wood blocks. When they either finished their design or the allotted time elapsed, they completed a survey assessing creativity self-efficacy and workload, as described in Section 3.6.

3) ***Revision Design Phase (30 min):*** After a 5 min break, participants refined their initial design by searching for precedent projects on ArchDaily for inspiration in the Control group or using DALL-E3 in the GenAI group. To ensure sufficient data collection, participants were asked to consult at least five architectural concepts to improve their design using ArchDaily or DALL-E3. After submitting their revised design, they completed the same survey assessing creativity self-efficacy and workload.

### 3.5 Design Performance Assessment

Participants' drawings of conceptual design were assessed by a panel of eight judges including four professors teaching design studios and four senior architectural students (fourth year undergraduate students enrolled in the accredited architectural engineering program at WPI). Prior to rating, the eight judges received a brief training session in which each rubric criterion was explained; however, this training did not include joint practice ratings, anchor-case discussion, or a consensus-building exercise, and inter-rater agreement was not re-assessed afterward. Each judge then independently assessed all 72 study drawings (36 participants × 2 phases), presented in randomized order on a purpose-built website. Each drawing was assessed on three dimensions using modified rubrics based on previous studies (Clary et al., 2011; Demirkan & Afacan, 2012; Gao et al., 2023) as described in Table 1. The rubric includes three core dimensions, clarity and legibility, complexity and detail, and overall visual communication. Each criterion was assessed using a 5-point Likert scale ranging from "1=Poor" to "5=Excellent." The design task performance was rated with the average of the three criteria. Inter-rater reliability for the composite rubric score, computed as the mean of the three rubric dimensions, was assessed using the average-measure intraclass correlation coefficient (ICC) across the eight raters. The average-measure ICCs were 0.157 for Phase 1 and 0.208 for Phase 2, indicating relatively low agreement among the eight raters. The three rubric dimensions were moderately to strongly intercorrelated in both phases (Phase 1: $r = 0.667–0.767$, all $ps < 0.001$; Phase 2: $r = 0.716–0.868$, all $ps < 0.001$), supporting their aggregation into a composite performance score.

***Table 1***. *Design performance evaluation rubrics*

| Dimension | 1 - Poor | 2 - Fair | 3 - Good | 4 -Very Good | 5 - Excellent |
|---|---|---|---|---|---|
| **Clarity and Legibility** | Sketches are unclear, with indistinct features or lines. | Sketches are somewhat clear but with inconsistent line quality. | Sketches are mostly clear, with minor readability issues. | Sketches are clear, legible, and understandable with minor issues. | Sketches are exceptionally clear, with all elements easily distinguishable. |
| **Complexity and Detail** | Minimal detail is included, with simplistic forms. | Some detail is present, but much of the design is too simple. | Adequate details are present but could be more comprehensive. | Design includes significant detail and demonstrates complexity. | Design is richly detailed, with all aspects clearly and intricately represented. |
| **Overall Visual Communication** | Design is poorly communicated visually, and it is difficult to understand the intent. | Visual communication is fair, but the design is hard to understand without further clarification. | Design is communicated well visually, with most aspects easy to understand. | Visual communication is very effective, with minor gaps. | Visual communication is highly effective, and the design is fully understandable with minimal effort. |

### 3.6 Self-efficacy in Creativity and Workload

After each design phase, participants completed questionnaires on creative self-efficacy and cognitive load. Creative self-efficacy was measured using a five-item self-report scale adapted from prior work (Pretz & McCollum, 2014), with each item rated on a 5-point Likert scale (1 = strongly disagree, 5 = strongly agree). As shown in Table 2, two indices were computed: task-specific creative self-efficacy, calculated as the mean of Q1 and Q2, and general creative self-efficacy, calculated as the mean of Q3–Q5. Cognitive load was assessed using the NASA Task Load Index (NASA-TLX) (Hart & Staveland, 1988), and the mean of the six subscales was used as the overall workload score (Table S2), following prior studies (Hart & Staveland, 1988; Devos et al., 2020; Frijns et al., 2024). Both creative self-efficacy subscales demonstrated good internal consistency: task-specific creative self-efficacy (Q1–Q2) yielded Cronbach's $\alpha = 0.816$ (Phase 1) and 0.869 (Phase 2); general creative self-efficacy (Q3–Q5) yielded $\alpha = 0.861$ (Phase 1) and 0.850 (Phase 2). The Raw NASA-TLX composite showed lower internal consistency ($\alpha = 0.243$ in Phase 1 and 0.442 in Phase 2.

***Table 2***. *Questions on self-efficacy in creativity*

| | Questions | Scale (1-5) |
|---|---|---|
| Task-specific creative self-efficacy | (Q1) My responses to the task were creative. | 1 = Strongly Disagree, 5 = Strongly Agree |
| | (Q2) My responses were more creative than the average person at my age. | 1 = Strongly Disagree, 5 = Strongly Agree |
| General creative self-efficacy | (Q3) I am good at coming up with new ideas. | 1 = Strongly Disagree, 5 = Strongly Agree |
| | (Q4) I have a lot of good ideas. | 1 = Strongly Disagree, 5 = Strongly Agree |
| | (Q5) I have a good imagination. | 1 = Strongly Disagree, 5 = Strongly Agree |

### 3.7 Interaction analysis

To examine how participants interacted with the assigned tool during the revision phase, we analyzed screen recordings from both the GenAI and Control conditions.

#### 3.7.1 Prompt analysis procedure

To relate prompt behavior to changes in cognitive load, we conducted a prompt analysis using prompts extracted from screen recordings of participants' GenAI interactions during the revision design phase. Participants generated their own prompts freely; no prompts were prescribed by the research team. This naturalistic design choice was deliberate: it allows the prompt analysis to capture the spontaneous interaction strategies that students develop when using GenAI without structured guidance, thereby providing ecologically valid data about how novice users engage with text-to-image AI in a design context. The absence of prompt training also reflects a realistic pedagogical scenario in which instructors have not yet explicitly taught prompt literacy.
A high quality prompt in this task context is one that is: (1) task-specific—tied to the design brief (research laboratory building massing and geometries); (2) iterative—building on the content or form of prior AI outputs rather than starting fresh each time; and (3) visually explicit— specifying formal, compositional, material, or spatial attributes that the AI can interpret and render. This definition is consistent with Urban et al. (2025), who characterize effective prompts in creative problem-solving as those that include explicit evaluation criteria, detailed contextual background, and iterative feedback. In contrast, vague or generic prompts (e.g., "give me a building") tend to produce outputs with low task relevance, requiring additional cognitive effort to evaluate and filter.
All prompts produced during the revision phase were coded using an eight-category scheme adapted from Urban et al. (2025), We originally used a nine-category scheme from that framework; however, the category "Providing Additional Context" was deleted after the expert review, as it did not manifest distinctively in our architectural task—participants who provided context invariably did so in the service of idea generation or visual feedback, making this category indistinguishable from CD2 or CD6 in our dataset. The final eight categories (CD1–CD8) are defined in Table 6 with examples. The first two authors and a senior undergraduate student in Architectural Engineering at WPI established labeling principles by jointly reviewing two sample cases; disagreements were resolved through discussion. The expert raters then independently coded the remaining prompts in randomized order, yielding acceptable to strong inter-rater reliability (weighted $\kappa = 0.73–0.78$) (Landis & Koch, 1977).

#### 3.7.2 Control search query extraction

For the Control condition, ArchDaily search queries were extracted from participants' screen recordings during the revision phase. Each participant entered search query was counted as one search interaction. Word count was calculated as the total number of words typed across all extracted search queries. These search-query measures were used only as behavioral evidence of tool engagement and were not part of the main prompt correlation analysis.

### 3.8 Statistical analysis

All analyses were conducted in Python using *scipy* and *statsmodels* (Seabold & Perktold, 2010; Virtanen et al., 2020). The analytical framework maps directly onto the three research questions:

RQ1 (overall effects): For between-group comparisons, we applied Shapiro-Wilk tests (selected for their superior power relative to Kolmogorov-Smirnov at $n < 50$ (Razali & Wah, 2011)) and Levene's tests to evaluate normality and homogeneity of variance, respectively. Depending on the results of these assumptions, Welch's independent-samples t-tests or Mann–Whitney U tests were applied for parametric dataset and non-parametric dataset, respectively. For within-group change across phases, outcomes were analyzed using difference-in-differences (DiD) regression:

$$Y_{it} = \beta_0 + \beta_1 \text{Tool}_i + \beta_2 \text{Post}_t + \beta_3 (\text{Tool}_i \times \text{Post}_t) + \varepsilon_{it} \quad (1)$$

where $\beta_3$captures the two phase change in the GenAI group relative to the control group. The outcome variable $Y_{it}$represents the score of participant $i$ at time $t$. The intercept $\beta_0$corresponds to the mean outcome of the control group during the initial design phase. The coefficient $\beta_1$captures the baseline difference between the GenAI and control groups prior to the intervention. The coefficient $\beta_2$represents the average change from the initial design phase to the revision design phase for the control group. The error term $\varepsilon_{it}$represents unexplained individual-level variation.

Standard errors were clustered by participant level to account for repeated measures. DiD estimates and their 95% confidence intervals were standardized by dividing them by the pooled standard deviation of the initial design phase scores.
RQ2 (competency moderation): initial design performance scores were dichotomized at the median into Novice and Experienced subgroups. This categorical approach is grounded in design cognition research (K.-L. Huang et al., 2024; Y. Ma et al., 2025), where differences in competency are often more meaningfully represented as distinct groups rather than as a purely continuous variable. Four two-way ANOVA models were conducted to test the interaction between Tool and Initial Competency Level across design performance, task-specific creative self-efficacy, general creative self-efficacy, and cognitive load. Because the Phase 1 performance ratings showed low inter-rater reliability, the subgroup analysis was interpreted as exploratory. Effect sizes were reported using partial eta-squared ($\eta^2_p$). All tests were two-tailed at $p = .05$.
RQ3 (prompt analysis): Delta cognitive load was calculated as the mean difference across six NASA-TLX items (revision phase – initial phase). Normality of distributions was assessed with the Shapiro-Wilk test. Pearson's r was used for approximately normal, linearly related variable pairs; Spearman's ρ was used where these assumptions were violated.
The study was not formally pre-registered. RQ1 and RQ2 were planned analyses based on the study's research questions, whereas RQ3 was treated as an exploratory, hypothesis-generating analysis of prompt-use patterns and cognitive-load change.

# 4. Results

## 4.1 Between-Group Comparisons across Different Design Phases

Because design performance was rated with low inter-rater reliability, the between-group and within-group comparisons involving this measure throughout Sections 4.1–4.3 should be read as attenuated in magnitude and inflated in uncertainty relative to what a more reliable measure would show. we return to a further, distinct consequence of this for the subgroup analysis in Section 4.3.
Table 3 presents the comparison of design performance, cognitive load, and creative self-efficacy between the Control group and the GenAI group across the Initial Design Phase and the Revision Design Phase. In the Initial Design Phase, participants in both groups developed their conceptual designs without the use of any external tool. No statistically significant differences were observed between the two groups for design performance, cognitive load, task-specific creative self-efficacy, or general creative self-efficacy (all p values > 0.32). The similarity in group means indicates that the random assignment procedure successfully produced comparable baseline conditions. Design performance ratings were approximately 2.8 to 2.9 on a five-point scale for both groups, and cognitive load remained lower than 3, suggesting a moderate level of perceived mental demand.
In the Revision Design Phase, the GenAI group demonstrated slightly higher design performance relative to the Control group (*Control: M = 2.785, SD = 0.361; GenAI: M = 2.903, SD = 0.433; p = 0.381)*, although this difference did not reach statistical significance. Cognitive load followed a similar pattern. Participants in the GenAI group reported a slightly lower level of cognitive demand than those in the Control group, but the difference was not significant. In contrast, both task-specific creative self-efficacy and general creative self-efficacy were modestly but not significantly lower in the GenAI group compared to the Control group.

***Table 3.*** *Comparison of design performance, cognitive load, and self-efficacy between Control Group and GenAI Group across two design phases.*

| **Design Phase** | **Tool Used** | **Parameter** | **Control (mean ± SD)** | **GenAI (mean ± SD)** | **P value** | **Cohen's *d*** |
|---|---|---|---|---|---|---|
| **Initial Design Phase** | None | Design performance | 2.921 ± 0.369 | 2.819 ± 0.388 | 0.426 | -0.269 |
| | | Cognitive load | 2.787 ± 0.322 | 2.778 ± 0.478 | 0.946 | -0.023 |
| | | Task-specific creative self-efficacy | 3.472 ± 0.866 | 3.556 ± 0.765 | 0.761 | 0.102 |
| | | General creative self-efficacy | 3.537 ± 0.76 | 3.796± 0.81 | 0.329 | 0.33 |
| **Revision Design Phase** | Control: ArchDaily<br>GenAI: DALL-E3 | Design performance | 2.785 ± 0.361 | 2.903 ± 0.433 | 0.381 | 0.296 |
| | | Cognitive load | 2.880 ± 0.442 | 2.843 ± 0.604 | 0.835 | -0.07 |
| | | Task-specific creative self-efficacy | 3.861 ± 0.854 | 3.556 ± 0.82 | 0.281 | -0.365 |
| | | General creative self-efficacy | 3.741 ± 0.805 | 3.574 ± 0.899 | 0.562 | -0.195 |

## 4.2 Difference in Difference Analysis across Design Phases

A difference-in-differences analysis was conducted to compare phase-to-phase changes between the GenAI-assisted and precedent-search workflows (Table 4, Fig. 4). The results indicate a positive estimated treatment effect for design performance (DiD = 0.583 SDs, 95% CI = [−0.476, 1.643]). This pattern reflects a numerically larger phase-to-phase improvement in the GenAI group relative to the Control group, driven by a slight increase in the GenAI group and a decline in the Control group; however, the confidence interval included zero. No significant treatment effect was observed for cognitive load (DiD = −0.069, 95% CI = [−0.742, 0.604]). To further unpack the workload results, we conducted an item-level DiD analysis for the six NASA-TLX dimensions (Fig. S1). This analysis showed a significant reduction in physical demand in the GenAI group relative to the control group (DiD = −0.55 SDs, 95% CI [−0.92, −0.17]), while frustration showed a marginal increase over time in the GenAI group (DiD = 0.76 SDs, 95% CI [−0.01, 1.52]).

In the control group, both task-specific creative self-efficacy ($p = 0.054$) and general creative self-efficacy ($p = 0.023$) exhibited similar upward trends from the initial to the revision design phase, indicating increased confidence following iterative design practice. In contrast, while the GenAI group showed a comparable pattern for task-specific creative self-efficacy, general creative self-efficacy declined during the revision design phase, resulting in a significant negative DiD effect (DiD = −0.543, 95% CI = [−1.068, −0.017]). As a result, the observed treatment effect appears to be driven primarily by the decrease in general creative self-efficacy in the GenAI group rather than by gains in the control group, suggesting a potential divergence between task-focused confidence and broader creative self-beliefs when GenAI support is introduced during design revision.

***Table 4.*** *DiD Estimates of Treatment Effects for GenAI Group*

| **Covariate** | **Initial Design Phase** | **Revision Design Phase** | **Initial Design Phase** | **Revision Design Phase** | **DID (SDs)** | **95%CI** |
|---|---|---|---|---|---|---|
| | ***Control*** | | ***GenAI*** | | | |
| **Design performance** | 2.921 ± 0.369 | 2.785 ± 0.361 | 2.819 ± 0.388 | 2.903 ± 0.433 | 0.583 | [-0.476, 1.643] |
| **Cognitive load** | 2.787 ± 0.322 | 2.880 ± 0.442 | 2.778 ± 0.478 | 2.843 ± 0.604 | -0.07 | [-0.742, 0.604] |
| **Task-specific creative self-efficacy** | 3.472 ± 0.866 | 3.861 ± 0.854 | 3.556 ± 0.765 | 3.556 ± 0.82 | -0.48 | [-1.233, 0.269] |
| **General creative self-efficacy** | 3.537 ± 0.76 | 3.741 ± 0.805 | 3.796± 0.81 | 3.574 ± 0.899 | -0.54 | [-1.068, -0.017] |

*DiD estimates and their 95% confidence intervals were standardized by dividing them by the pooled standard deviation of the initial design phase scores.*

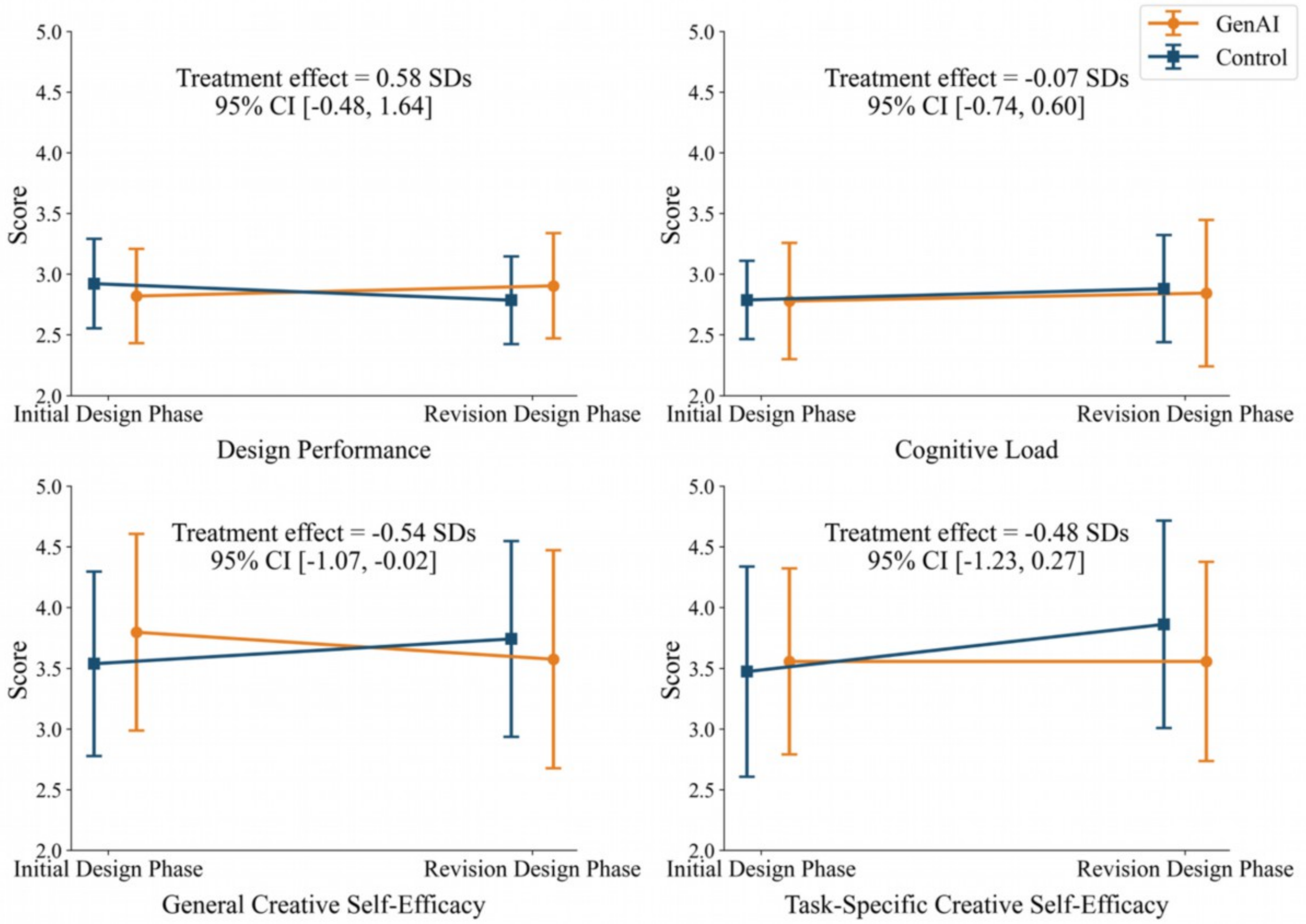

***Fig. 4.*** *Within-Group Changes in Design Performance from the Initial Phase to the Revision Phase for the GenAI and Control Groups. Displayed are the treatment effect coefficients and their corresponding 95% confidence intervals, rescaled in units of the outcome variable's standard deviation from the initial design phase.*

### 4.3 Examining the Impact of Initial Competency Level on Outcomes in Revision Design Phase

A two-way ANOVA revealed a significant interaction effect between tool condition and initial competency level on design performance during the Revision Phase *($F(1, 32) = 4.303$, $p = 0.046$, partial $\eta^2 = 0.118$)* (Table 5). As illustrated in Fig. 5, participants classified as Novice who used GenAI achieved higher revision-phase design-performance ratings than those in the Control condition (*$p = 0.023$*). No corresponding difference was observed between the GenAI and Control conditions among participants classified as Experienced (*$p = 0.449$*).
This interaction should be interpreted cautiously. The Novice and Experienced groups were created by median-splitting the Phase 1 performance ratings, which showed low inter-rater reliability. Measurement uncertainty in these ratings may therefore have affected subgroup assignment. Together with the small subgroup cell sizes, the wide 95% confidence interval for partial $\eta_p^2$, [0.000, 0.317], and the limited observed power of 0.569, this limits the stability of the interaction. The finding is therefore presented as an exploratory pattern rather than as robust evidence of a competency-dependent effect. No significant main effects or interactions were identified for the remaining outcomes (Table S3).

***Table 5.*** *Two-way ANOVA results for the effects of initial competency level on design performance during the revision phase*

| **Variable** | **SS** | **df** | **F** | **Sig.** | **Partial $\eta^2$** |
|---|---|---|---|---|---|
| Tool | 0.106 | 1 | 0.745 | 0.394 | 0.022 |
| Initial Competency Level | 0.263 | 1 | 1.857 | 0.182 | 0.055 |
| Tool* Initial Competency Level | 0.61 | 1 | 4.303 | **0.046** | 0.118 |

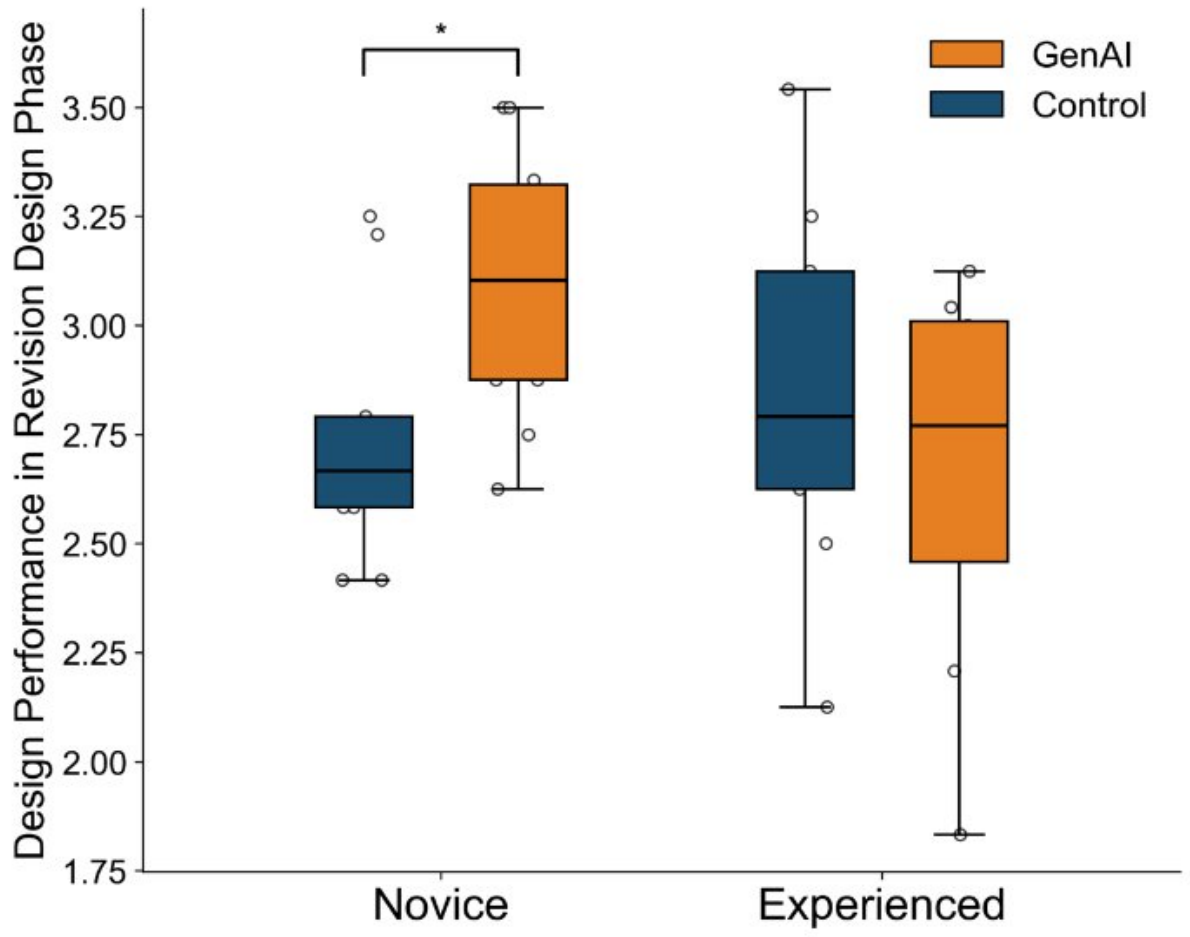

*Fig. 5. The effects of initial competency level on design performance during the revision phase. *: p < 0.05.*

## 4.4 Associations between GenAI prompt use and cognitive load change

We identified eight prompt categories in our coding scheme. Sixteen participants who used GenAI were included in the analysis, with two excluded due to missing screen-recording data. Across the 16 participants who used GenAI prompts, prompt usage varied widely in both quantity and type. Participants produced between 3 and 26 prompts (M = 9.38 ± 6.02) with word counts ranging from 20 to 295 words (M = 110 ± 74.73). Neither prompt count nor word count were significantly associated with changes in cognitive load (both *ps > .05*) as described in Table 6.

Among the eight prompt categories, CD2 (Idea generation, specifically for visual communication) was used most frequently (M = 4.38 ± 4.33), followed by CD3 (Idea generation, based on previous answer). Correlation analyses revealed two significant negative relationships: greater use of CD3 prompts (*r = –0.566, p = 0.022*) and CD6 prompts (*r = –0.518, p = 0.039*) were associated with larger decreases in cognitive load. CD3 and CD6 represent complementary phases of an iterative refinement cycle: CD3 is a forward-generative strategy (selecting a prior output and requesting further development), while CD6 is a backward-evaluative strategy (directing the AI to revise a specific aspect relative to a quality criterion). Together, these two categories reflect the generate–evaluate–refine cycle that characterizes expert design cognition. The associations observed for CD3 and CD6 suggest that prompts supporting iterative refinement and visual communication may be linked to reduced cognitive load during design revision. Although these associations were significant at the uncorrected level but did not survive correction for multiple comparisons: CD3, Bonferroni-adjusted $p = 0.177$, BH-FDR-adjusted $p = 0.133$; CD6, Bonferroni-adjusted $p = 0.265$, BH-FDR-adjusted $p = 0.133$. Therefore, these findings should be interpreted as exploratory rather than confirmatory effects.

### 4.5 Tool Interaction Patterns and Behavioral Engagement Across Conditions

Screen recordings from the Control condition (N = 17; one participant excluded due to missing data) yielded ArchDaily search queries that were compared with GenAI prompt data (N = 16). GenAI participants entered significantly more prompt interactions than Control participants (GenAI: M = 9.38 ± 6.02 prompts; Control: M = 4.71 ± 4.83 search queries; $p = 0.021$). Text-based interaction differed more substantially (GenAI: M = 110 ± 74.73 words; Control: M = 9.24 ± 9.63 words; $p < 0.001$), with Control queries largely consisting of brief and repeated keywords (e.g., one participant entered "arches" three times within the same session) rather than the descriptive, contextually rich language characteristic of GenAI prompting. Twelve of 17 Control participants entered at least one query directly referencing the building program (e.g., "research lab," "research lab façade," "academic building"). Several demonstrated iterative search refinement within their session: for example, one participant progressively narrowed from broad structural queries ("arches," "windows") to specific visual attributes ("bubble window," "stained glass designs"). In the GenAI condition, one participant entered "green roof" as a prompt and subsequently added green vegetation elements to the revised design.

***Table 6.*** *Descriptive statistics and intercorrelations between prompts and changes of cognitive load, qualitative description, and examples of individual prompts.*

| Code | Variable | Min | Max | M ± SD | Correlation coefficient[†] *(uncorrected p value)* | Description | Example |
|---|---|---|---|---|---|---|---|
| | Prompt count | 3 | 26 | 9.38 ± 6.02 | -0.261 (>0.05) | Range of prompt counts | |
| | Word count | 20 | 295 | 110 ± 74.73 | -0.003 (>0.05) | Range of word counts | |
| **CD1** | Generation (based on task instruction) | 0 | 4 | 1.06 ± 1.34 | 0.075 (>0.05) | The initial prompt rephrasing information from a task instruction. | *"Draw a building."* *"Give me 10 images of modern laboratory buildings."* |
| **CD2** | Idea generation (specifically for visual communication) | 0 | 14 | 4.38 ± 4.33 | -0.312 (>0.05) | Specific instruction for specific idea generation evaluation criteria (effective communication and understandable). | *"Can you please generate me two more buildings in the shape of the letter O and another in the shape of S, both buildings should have a brick exterior."* *"Add a courtyard."* |
| **CD3** | Idea generation (based on previous answer) | 0 | 5 | 1.06 ± 1.39 | -0.566 (0.022) | Participants selected one idea from previous answers and asked GenAI to generate more ideas based on their selection. | *"Now make the building shorter, while maintaining the gabled roof."* *"Make 5 more."* |
| **CD4** | Information gathering | 0 | 8 | 0.75 ± 2.02 | 0.021 (>0.05) | Using GenAI as an information resource by creating image | *"Head of snake building."* *"Show me a modern laboratory."* |
| **CD5** | Feedback (general) | 0 | 1 | 0.19 ± 0.40 | 0.035 (>0.05) | Feedback without any specific details. | *"Good job."* |
| **CD6** | Feedback (improve visual communication) | 0 | 12 | 1.63 ± 3.12 | -0.518 (0.039) | Feedback specifically asking GenAI to improve communication and understand understandability | *"Too spiky, try again."* *"This picture also has grass on the roof, you failed. Try again."* |
| **CD7** | Asking for feedback | 0 | 1 | 0.13 ± 0.34 | 0.041 (>0.05) | Asking GenAI to provide feedback on ideas that were created. | *"What might such a building look like?"* |
| **CD8** | Regenerate | 0 | 2 | 0.19 ± 0.54 | 0.391 (>0.05) | Pressing the "regenerate" button in the chat console. | ⟳ |

*Min represents the minimal value obtained by an individual participant. Max represents the maximal value obtained by an individual participant. M and SD represent the mean and standard deviation values obtained by an individual participant, respectively.*

*†: Spearman's r coefficient is reported for association between prompt features and delta cognitive load. This table follows the framework proposed by Urban et al. (2025).*

## 5. Discussion

This study compared design performance, creative self-efficacy, and perceived workload across a GenAI-assisted image-generation workflow and an ArchDaily-based precedent-search workflow. These patterns reflect the comparison between two bundled workflows that differed jointly in tool function, content source, and interaction mode. They therefore cannot be attributed to any single component in isolation. The mechanisms discussed below should therefore be understood as theoretically informed interpretations rather than component-specific causal tests.

### 5.1 GenAI workflow may support early-stage ideation for novice students

In the traditional architectural conceptual design workflow, designers typically begin with broad brainstorming followed by in-depth site analysis to produce preliminary concept sketches (Filzmoser et al., 2017). The time-consuming nature of the work significantly constrains creativity and exploration, especially for students who are still developing the ability to externalize and evaluate design ideas efficiently (Wang et al., 2025). Our subgroup analysis suggests that GenAI may scaffold this stage of design particularly for students with lower initial competency, helping them translate abstract intentions into inspectable visual alternatives. These results align with observations reported in prior studies (K.-L. Huang et al., 2024; Kim & Lee, 2023). However, because this grouping was derived by median-splitting a performance measure with low inter-rater reliability of 8 raters, we treat this as a preliminary pattern rather than evidence of a reliable novice-specific benefit, and the discussion below should be read in that light.

Unlike experienced designers who can fluently generate, iterate, and evaluate spatial ideas through sketching and analog reasoning (Goldschmidt, 2014), novices lack this cognitive infrastructure (Thurlow & Ford, 2018). Our subgroup analysis suggests that GenAI workflow may scaffold this early stage of design particularly for students with lower initial competency by helping them translate abstract intentions into inspectable visual alternatives. These results align with prior studies showing that GenAI can support ideation and visual exploration in architectural education (Wang et al., 2025), partially compensating for novices' limited ability to specify visual ideas precisely (Hollan et al., 2000; Horsky et al., 2003). The iterative refinement workflow within a single conversational session also allows students to generate an image, inspect its relevance to their intention, and revise prompts while retaining prior context. In this way, the system may support a generate–evaluate–refine cycle that helps students progressively clarify underdeveloped design concepts (Antony & Huang, 2024; L. Li & Bertrand, 2026).

At the same time, this potential scaffolding appears to support early visual ideation more than deeper architectural resolution (C. Li et al., 2025). Prior review suggest that generative AI in architecture remains largely image-driven, with evaluation often privileging aesthetic richness over spatial, and functional performance (S. Jang et al., 2025). Because architectural design depends on tightly interdependent relationships among concept, form, plan, structure, and section, current text-to-image systems are better suited to generating concept images and stylistic alternatives than to supporting reasoning about feasibility, spatial organization, and overall coherence (Liu et al., 2026; N. Ma et al., 2026). This may help explain why no comparable benefit was observed among experienced students, which is consistent with prior studies in other creative tasks (M. Huang et al., 2024).

Experienced students often possess more refined mental models, stronger internal evaluation criteria, and more stable strategies for generating and evaluating design alternatives (K.-L. Huang et al., 2024), making support that is primarily visual and idea-externalizing of limited additional value in short time exposure. For these individuals, AI outputs that are generic or architecturally imprecise may function as ideational interference rather than inspiration, disrupting established design thinking rather than enriching it (Y. Ma et al., 2025). This exploratory pattern is broadly consistent with the asymmetrical enhancement proposed by Li et al., wherein creative facilitation for novices coexists with constrained or unchanged performance for advanced learners (C. Li et al., 2026). Recent meta-analytic evidence further indicates that the effects of ChatGPT are significantly moderated by intervention setting, with laboratory-based designs yielding non-significant effects compared to classroom implementations (Deng et al., 2025). This contextual difference may also partly explain the absence of broad performance gains in our study.

### 5.2 GenAI affects creative self-evaluation in human–AI visual collaboration

Unlike findings in writing-based tasks where GenAI use is often associated with self-efficacy gains (Pellas, 2023), our visual communication task produced no significant improvement in task-specific creative self-efficacy, consistent with recent meta-analyses. Moreover, the GenAI group showed a significant decline in general creative self-efficacy (DiD = $-0.543$, $p < 0.05$). This finding, alongside the exploratory pattern of improved design performance among novice students, may suggest a potential decoupling between task support and self-belief that warrants careful pedagogical attention. On a technical level, this stagnation may be attributable to the limitations of early GenAI models in producing specific architectural concepts or accurately interpreting prompts (Alon et al., 2026); such friction can directly undermine a student's perceived competence during the task. The observed marginal increase in frustration (DiD = 0.76 SDs) further supports the interpretation that technical friction contributed to affective disengagement during the task. However, technical limitations alone do not explain the potential disconnect suggested by the exploratory novice pattern. This possible divergence may help explain why improved task support did not translate into stronger general creative self-efficacy. Although GenAI can temporarily compensate for practical barriers in idea development, it does not automatically strengthen students' broader self-beliefs as creative agents (S. Li et al., 2025). This phenomenon may reflect how GenAI reshapes the evaluative frame through which learners assess their own creativity. When working alongside a highly capable generative system, learners' internal standards for originality and creativity may shift upward, resulting in more conservative self-assessments even when design outcomes are comparable (Festinger, 1954).

From a design cognition perspective, creative self-efficacy is closely tied to perceptions of authorship and creative agency (Tierney & Farmer, 2002). Working with GenAI may blur the boundary between self- and system-generated contributions by redistributing cognitive effort away from idea generation toward the evaluation and curation of AI outputs. Students who cannot clearly attribute design decisions to themselves may experience a diminished sense of creative ownership, consistent with Glaveanu 's (2014) distributed creativity framework, wherein creative agency becomes perceived as a property of the human–AI system rather than of the individual. This distributed attribution may paradoxically weaken individual creative self-beliefs even as it strengthens task outcomes. These findings underscore the critical need for pedagogical framing that positions GenAI as a subordinate tool requiring human curation, thereby reinforcing the learner's agency and preserving confidence in their distinct creative role.

### 5.3 Prompt strategies regulate cognitive burden in GenAI-supported design

Despite the exploratory pattern of higher revision-phase design performance among novice GenAI users, no overall reduction in cognitive load was observed for GenAI users relative to the control group, consistent with recent research on AI integration in conceptual design workflows (Fu et al., 2026). In terms of tool-interaction behavior, the two conditions differed substantially in textual input demands. GenAI participants produced significantly more elaborate inputs than Control participants, suggesting that composing effective prompts itself constitutes a non-trivial cognitive task that may offset part of the ideational load relief GenAI provides. The further prompt analysis reveals that greater use of CD3 prompts (iterative idea generation based on prior outputs) and CD6 prompts (feedback targeting visual communication improvement) were each significantly associated with larger decreases in cognitive load. Crucially, total prompt count and word count showed no such association, confirming that volume of interaction is not what matters, but rather the nature of engagement. Rather than uniformly simplifying the design process, GenAI appeared to redistribute cognitive effort from ideation toward interaction management, prompt formulation, output evaluation, and iterative refinement (Bozkurt, 2024; Walter, 2024). As Zamfirescu-Pereira et al. (2023) demonstrate, non-AI experts frequently struggle to formulate effective prompts, particularly in creative domains where semantics, style, and spatial logic are tightly coupled. This prompting challenge constitutes a new form of extraneous cognitive load that may offset the intrinsic load reduction afforded by AI-assisted idea generation. Prior studies indicate that simplified or imprecise prompts often fail to elicit high-quality AI responses, particularly in creative domains (Urban et al., 2025; Wadinambiarachchi et al., 2024; Yilmaz & Karaoglan Yilmaz, 2023; Zamfirescu-Pereira et al., 2023), and our findings extend this evidence to the architectural visual design context.

CD3 and CD6 together reflect a generate–evaluate–refine cycle: CD3 is a forward-generative strategy (selecting a prior AI output and requesting further development), while CD6 is a backward-evaluative strategy (directing the AI to revise a specific aspect against a design criterion). Conceptually, these prompt types may help narrow the design space and clarify task-relevant information, thereby reducing some of the ambiguity that contributes to cognitive load. However, this interpretation should be treated as exploratory because the prompt-category associations were based on a small GenAI subsample and did not survive correction for multiple comparisons.

By contrast, initial-generation prompts such as CD1 and CD2 did not show comparable associations with cognitive-load reduction. This pattern is consistent with evidence that passive or undifferentiated AI use can sustain or increase cognitive burden (Bozkurt, 2024; Y. Chen et al., 2025; Fan et al., 2025; Subramonyam et al., 2024). Notably, the combined CD3/CD6 strategy shares structural similarities with the hybrid feedback approach identified by Alsaiari et al. (2026), where combining directive and reflective elements produced the deepest student revision behavior, suggesting that iterative, evaluative engagement with AI is a generalizable mechanism across different educational AI interactions.

To summarize, these findings contribute to ongoing discussions of the productivity paradox of emerging technologies: while we did not find cognitive load reduction for the GenAI group in general, the improper prompting can even increase the cognitive load due to learning curves, interface friction, and limited domain alignment (Zamfirescu-Pereira et al., 2023). In architectural design education, GenAI's potential benefits therefore remain inseparable from the need for structured guidance and pedagogical support.

### 5.4 Educational Innovation

This study makes a methodological contribution by applying a pedagogical data analytics lens to GenAI in design education. Rather than focusing only on final design outcomes, we integrate four levels of analysis: group-level performance and self-reports, within-group change through a DiD model, an exploratory competency subgroup analysis through a two-way ANOVA model, and individual prompt behavior from screen recordings, to capture how students interact with, respond to, and learn with GenAI during design. In addition, by extracting and coding naturalistic prompts without prescribing interaction patterns, we provide an ecologically valid method for examining actual GenAI use in creative tasks. The resulting eight-category coding scheme offers a reusable framework for future studies of human–AI interaction in design education.

**5.5 Limitations and Future Work**

This study has several limitations. First, with 36 participants further divided into subgroups, the small cell sizes limit the precision and generalizability of the Tool × Initial Competency Level interaction. Although the interaction was statistically significant, its effect-size confidence interval was wide and observed power was limited. The subgroup finding should therefore be interpreted as exploratory rather than definitive. Similarly, the prompt-category analyses were based on a small GenAI subsample, and the CD3 and CD6 associations were significant only at the uncorrected level. These associations should be treated as hypothesis-generating and require confirmation in larger samples with prespecified hypotheses. Second, inter-rater reliability for the design-performance ratings was relatively low, which introduces measurement uncertainty into the performance outcomes. One potential explanation is that creative design tasks rarely have a single “correct” answer and are therefore susceptible to substantial subjectivity in evaluation. Similar reliability challenges have been reported in other creative domains, including symbolic music generation (Ghoshal et al., 2026) and argumentative essay writing (Vögelin et al., 2019). This limitation is particularly relevant to the RQ2 analysis because the Novice and Experienced groups were created by median-splitting the same Phase 1 ratings. Measurement uncertainty may therefore have affected subgroup assignment, while median dichotomization further reduced statistical precision. A continuous moderation approach in a larger sample would provide more robust estimates of competency-level differences. Future studies should also incorporate more systematic rater preparation, including pre-rating training, calibration meetings, and anchor-based practice, to improve consistency in the evaluation of open-ended conceptual design work. Third, the 30-minute revision phase is substantially shorter than real-world design workflows, which may have constrained participants' prompting strategies and underestimated the cognitive benefits of iterative AI use in naturalistic settings. Fourth, this study examined a single GenAI system, DALL-E 3, within the technological context available at the time of the experiment. Therefore, the findings should be understood as a time-bounded evaluation of one accessible platform rather than a general benchmark for AI-supported architectural design. Finally, Prior AI experience was not formally assessed. And neither condition included a post-task self-report item assessing the extent to which participants felt their assigned tool influenced their revised design, nor were post-task interviews conducted. Both would have strengthened the manipulation check and provided insight into participants' sensemaking and adaptation strategies; future studies should incorporate both. Future work should adopt longitudinal and mixed-methods approaches to examine how human–AI collaboration, prompt strategies, and design competencies evolve over time, and to inform responsible and pedagogically grounded integration of GenAI in architectural education.

## 6. Conclusions

This study compared a GenAI-assisted image-generation workflow with an ArchDaily-based precedent-search workflow in an architectural conceptual design task. Accordingly, the contribution of this study is evidence about these complete workflows rather than an estimate of the isolated effect of GenAI itself. The two workflows showed no significant overall differences in design performance, cognitive workload, or task-specific creative self-efficacy, while general creative self-efficacy declined relatively under the GenAI workflow. Exploratory analyses suggested a possible novice performance benefit. The GenAI workflow also involved substantially greater interaction volume and verbal elaboration than the precedent-search workflow, highlighting the interaction behavior differences between the two conditions. Iterative and task-specific prompt-use patterns were associated with cognitive workload reduction at the uncorrected level.

These findings suggest three cautious implications for the educational integration of GenAI in visual-spatial design disciplines. First, our findings are consistent with the possibility that GenAI may function as a conditional scaffold rather than a universal enhancement, with potential benefits among students with lower initial design performance and outcomes that may vary with human–AI interaction patterns. This pattern warrants further investigation in larger samples and across additional design contexts. Pedagogically, the findings suggest that instructors may consider adapting AI scaffolding to students' initial competency, introducing GenAI in workflow as an ideation support for novices while positioning it as a critical evaluation tool for more experienced learners. Second, the decline in general creative self-efficacy underscores the need for deliberate pedagogical framing that preserves students' sense of creative agency. Sequencing AI-mediated phases with AI-free phases, and explicitly asking students to distinguish their own contributions from AI-generated content, may help sustain creative self-beliefs during GenAI-supported design activities. Third, prompt literacy may represent a teachable competency worthy of further investigation. Students who engaged in task-specific, iterative prompting showed greater cognitive workload reductions at the uncorrected level, although these associations remain exploratory. GenAI-integrated design curricula may therefore consider explicitly supporting prompt formulation alongside traditional visual communication skills, rather than assuming that students already possess this capability.

Taken as a whole, these results may indicate that the educational value of GenAI lies not in uniformly enhancing outcomes in visual-spatial design and other complex creative tasks, but may vary according to learner characteristics, pedagogical framing, and interaction practices. Future implementation may therefore benefit from ability-sensitive support, deliberate protection of creative agency, and structured development of human–AI interaction skills.

## Acknowledgements

This research was supported by U.S. National Science Foundation (#2417102). Any opinions, findings, conclusions or recommendations expressed in this work are those of the authors and do not necessarily reflect the views of the National Science Foundation. We would like to thank Prof. Nan Ma for her helpful feedback on the experimental design.

**Author contributions**
Yao Xiao: Writing – original draft, Visualization, Methodology, Investigation, Formal analysis, Revision. Han Jiang: Writing – original draft, Visualization, Investigation, Preliminary analysis, Conceptualization. Rachel Hurley: Data acquisition. Shichao Liu: Writing - review & editing, Supervision, Funding acquisition.

**Declaration of Generative AI and AI-assisted technologies in the writing process**
During the preparation of this work the authors used Gemini and ChatGPT in order to improve grammar, spelling and clarity. After using this tool/service, the authors reviewed and edited the content as needed and take full responsibility for the content of the publication.

**Declaration of competing interest**
The authors declare that they have no known competing financial interests or personal relationships that could have appeared to influence the work reported in this paper.

**Statement on open data and ethics**
This study was approved by the Institutional Review Board (IRB) at Worcester Polytechnic Institute (WPI) (IRB-20–0001). Informed consent was obtained from all participants, and their privacy rights were strictly observed. The data supporting the findings of this study are available on the Open Science Framework (OSF):
https://osf.io/b73jp/overview?view_only=467c635b01cb40cc98f3fcb7aeffe0a8